%% file: linkerllm_weightsym.tex
\documentclass{article}

\usepackage{microtype}
\usepackage{graphicx}
\usepackage{booktabs}
\usepackage{hyperref}
\usepackage[preprint]{icml2026}
\usepackage{amsmath}
\usepackage{amssymb}
\usepackage{xcolor}
\usepackage{multirow}
\usepackage{algorithm}
\usepackage{algorithmic}
\usepackage{subcaption}

\icmltitlerunning{Block-Level Weight-Space Structure Under Post-Training}

\begin{document}

\twocolumn[
  \icmltitle{Block-Level Weight-Space Structure Persists Under \\Post-Training: An Empirical Study Across LLM Families}

  \icmlsetsymbol{equal}{*}

  \begin{icmlauthorlist}
  \icmlauthor{Zhaohui Wang}{usc}
  \end{icmlauthorlist}

  \icmlaffiliation{usc}{Viterbi School of Engineering, University of Southern California, Los Angeles, USA}
  \icmlcorrespondingauthor{Zhaohui Wang}{zwang000@usc.edu}

  \vskip 0.3in
]

\printAffiliationsAndNotice{}

\begin{abstract}
Modern LLMs are deployed as families of post-trained variants (base, instruct, chat, code) derived from a shared set of pre-trained weights. We present an empirical study of how post-training transforms weight-space geometry, covering eight configurations across four architecture families (Qwen2.5, Llama-3.1/3.2, Mistral, Gemma-2). We identify a \emph{granularity gap}: post-training modifies every tensor (zero of 291--339 tensors remain byte-identical, so hash-based deduplication achieves 0\% savings), yet preserves block-level structure (mean cosine similarity exceeds 0.99 and relative Frobenius distance stays below 0.13). Post-training therefore acts as a \emph{structured perturbation} that shifts every parameter while leaving block-level geometry intact. The property is not universal: independently trained specializations (e.g., Qwen2.5-Coder) attain cosine similarity $\sim$0.64 with the general base, indicating a disconnected region of weight space. Perturbation magnitude varies systematically with model scale, architecture family, and post-training recipe. As a practical application, we build \textbf{LinkerLLM}, a lazy loader that aliases shareable blocks across co-resident variants, achieving 18--48\% GPU memory savings and enabling up to five 7B-parameter variants on a single 24\,GB consumer GPU. Five of eight configurations retain $\ge$94\% of the unshared variant's quality on MMLU, ARC-Challenge, HellaSwag, and WinoGrande; the remaining three (Mistral-7B, Gemma-2-2B, Llama-3.2-1B) have one below-threshold benchmark each (87--91\%), which we report transparently rather than gate the block-sharing decision on a single threshold.
\end{abstract}

\section{Introduction}
\label{sec:intro}

How do training procedures transform the geometry of neural network weights? Linear mode connectivity~\citep{frankle2020linear,entezari2022role} established that models fine-tuned from the same initialization remain connected by low-loss paths, and Git Re-Basin~\citep{ainsworth2023git} extended this to independently trained models via permutation alignment. We study a specific, practically motivated instance of this question: \emph{what happens to weight-space structure when a pre-trained LLM undergoes post-training?} Post-training (SFT, RLHF, DPO~\citep{grattafiori2024llama3,yang2024qwen2}) produces specialized variants (instruct, chat, code) sharing a common origin.

Our study reveals a previously undocumented \textbf{granularity gap}: post-training perturbs the weight space at the tensor level (every tensor changes; hash-based deduplication is useless) while preserving structure at the block level (cosine $>$0.99, Frobenius ratio $<$0.16 across all eight tested configurations). The property is not universal: independently trained specializations such as Qwen2.5-Coder-7B fall outside the block-connected neighborhood (cos $\sim$0.64). We further verify that block-level alignment is \emph{not} a consequence of the architecture's permutation symmetry: a function-preserving permutation drops the cosine by three orders of magnitude. As a practical application, we build \textbf{LinkerLLM}, a runtime loader that aliases shareable blocks across co-resident variants, achieving 18--48\% GPU memory savings on 7B-class models on a 24\,GB consumer GPU.

\textbf{Related work.} The mode connectivity literature characterizes basins at the whole-model level~\citep{frankle2020linear,entezari2022role,ainsworth2023git,theus2025glmc}; our analysis refines the picture at the \emph{block} level. Task-vector approaches~\citep{ilharco2023editing,yadav2023ties,yu2024dare,wortsman2022model,sun2025amurochar} operate on weight-space \emph{deltas}; we characterize the \emph{structure} of those deltas. Multi-model serving systems share LoRA adapters~\citep{sheng2023slora,chen2023punica,hu2021lora} or compress deltas~\citep{yao2025deltazip,wang2025zipllm}; our empirical finding explains why tensor-hash dedup fails and motivates block-level sharing instead. Concurrent work~\citep{zhong2025watchweights} reads structure off fine-tuned weights for monitoring and control; our focus is structural quantification and memory-efficient deployment.

\section{Methodology}
\label{sec:method}

For a transformer block $i$ with sub-parameters $\{W^{(i)}_k\}_{k=1}^K$ (attention projections, MLP weights, layernorms), we define two complementary metrics between a base model and a post-trained variant:
\begin{equation}
\text{sim}(i) = \frac{1}{K}\sum_{k=1}^{K} \frac{\langle \text{vec}(W^{(i)}_{k,\text{base}}),\; \text{vec}(W^{(i)}_{k,\text{var}}) \rangle}{\|\text{vec}(W^{(i)}_{k,\text{base}})\| \cdot \|\text{vec}(W^{(i)}_{k,\text{var}})\|}
\label{eq:cosine}
\end{equation}
\begin{equation}
\epsilon(i) = \frac{1}{K}\sum_{k=1}^{K} \frac{\|W^{(i)}_{k,\text{base}} - W^{(i)}_{k,\text{var}}\|_F}{\|W^{(i)}_{k,\text{base}}\|_F}
\label{eq:frobenius}
\end{equation}
Cosine measures \emph{directional} preservation; Frobenius ratio measures \emph{magnitude} of perturbation. Both are needed: a block can have high cosine but large Frobenius (e.g., uniform scaling), or vice versa. A block is \emph{shareable} when both pass per-family thresholds:
\begin{equation}
\text{Share}(i) \iff \text{sim}(i) \geq \tau_\text{cos} \;\wedge\; \epsilon(i) \leq \tau_\text{frob}
\label{eq:criterion}
\end{equation}
We study eight base$\to$instruct pairs across four architecture families (Qwen2.5-0.5/3/7B, Llama-3.2-1/3B, Llama-3.1-8B, Mistral-7B-v0.3, Gemma-2-2B). Weights are compared in float32 after conversion from the model's native precision; full configuration details appear in Appendix~\ref{app:models}.

\section{Results}
\label{sec:results}

\subsection{The Granularity Gap}

\textbf{Tensor level.} For Qwen2.5-7B base vs.\ instruct, only 2 of 339 tensors are byte-identical (both small bias vectors); for Llama-3.1-8B, zero of 291 match. Post-training produces a \emph{dense} perturbation, with no tensor left untouched.

\begin{figure}[t]
\centering
\includegraphics[width=\columnwidth]{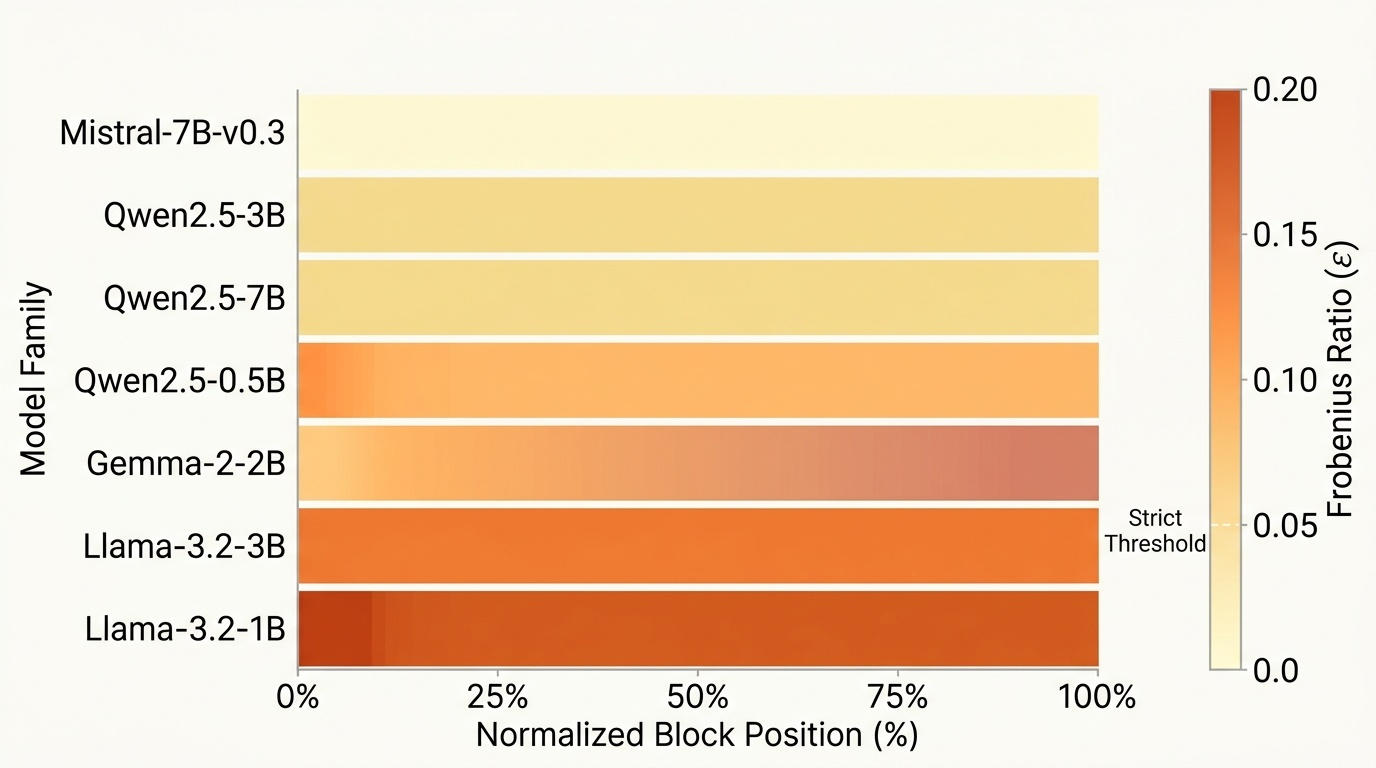}
\caption{Per-block Frobenius ratio $\epsilon(i)$ across model families (base$\to$instruct). Mistral and Qwen families have uniformly low perturbation (all blocks near-frozen). The Llama-3.2 series shows 10--20$\times$ larger per-block perturbation than Mistral, consistent with more aggressive RLHF; the Llama-3.1-8B variant lies between these regimes ($\bar\epsilon=0.045$), suggesting that perturbation magnitude is not a fixed property of the Llama family but varies with the specific post-training recipe used per release.}
\label{fig:frob}
\end{figure}

\textbf{Block level.} Despite per-tensor changes, transformer blocks retain high similarity (Table~\ref{tab:similarity}, Figure~\ref{fig:frob}): cosine $>$0.99 across all eight configurations and Frobenius ratio $<$0.05 for six of eight families. The perturbation is ``block-structured'': it modifies every tensor within a block by a small, correlated amount that preserves the block's function. The block-level mean is representative: per-sub-parameter cosines on Qwen2.5-7B block~14 range from 0.9997 to 1.0000 with std.\ 0.009 (Appendix~\ref{app:max-over-k}).

\begin{table}[t]
\caption{Block-level similarity between base and instruct variants. Families ordered by mean Frobenius ratio (ascending = more preserved).}
\label{tab:similarity}
\vskip 0.1in
\centering
\small
\setlength{\tabcolsep}{4pt}
\begin{tabular}{@{}lcccc@{}}
\toprule
Family & Blocks & cos range & $\epsilon$ range & $\epsilon$ mean \\
\midrule
Mistral-7B-v0.3 & 32 & 0.999+ & 0.004--0.01 & 0.006 \\
Qwen2.5-3B & 36 & 0.999+ & 0.01--0.02 & 0.011 \\
Qwen2.5-7B & 28 & 0.999+ & 0.01--0.02 & 0.015 \\
Llama-3.1-8B & 32 & 0.999+ & 0.04--0.05 & 0.045 \\
Qwen2.5-0.5B & 24 & 0.998 & 0.04--0.05 & 0.045 \\
Gemma-2-2B & 26 & 0.998--0.999 & 0.04--0.08 & 0.049 \\
Llama-3.2-3B & 28 & 0.991--0.995 & 0.10--0.12 & 0.110 \\
Llama-3.2-1B & 16 & 0.990--0.995 & 0.12--0.17 & 0.156 \\
\bottomrule
\end{tabular}
\end{table}

\textbf{Correlates of perturbation magnitude.} The mean Frobenius ratio $\bar\epsilon$ varies with (i) model scale (sub-1B Qwen models receive 3--4$\times$ larger perturbation than $\ge$3B), (ii) training methodology (predominantly-SFT Mistral-v0.3 has $\bar\epsilon{=}0.006$ while heavier-RLHF Llama-3.2 reaches $0.110$--$0.156$), and (iii) release vintage (Mistral-v0.1$\to$v0.3 and Gemma-1$\to$Gemma-2 show order-of-magnitude tightening between 2023 and 2024 releases, though Llama-2$\to$Llama-3.1 is essentially flat). These correlations are confounded by simultaneous differences in architecture and data; controlled studies would be needed to disentangle them. Full analysis: Appendix~\ref{app:version-trend}.

\begin{figure}[t]
\centering
\includegraphics[width=0.95\columnwidth]{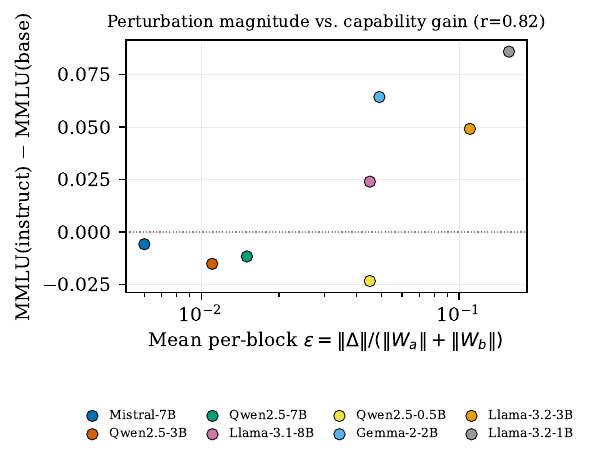}
\caption{Per-family perturbation magnitude $\bar\epsilon$ vs.\ MMLU capability gain (Pearson $r=0.819$, $n=8$). Mistral-v0.3 / Qwen-3B / Qwen-7B with $\bar\epsilon\in[0.006, 0.015]$ show flat or slightly negative MMLU change; larger-$\bar\epsilon$ families (Llama-3.2 series, Gemma-2-2B) show 5--9\,pt gain. $n=30$ per MMLU subject (lm-eval v0.4.11).}
\label{fig:eps-vs-gain}
\end{figure}

\textbf{Perturbation magnitude vs.\ downstream capability gain.} Figure~\ref{fig:eps-vs-gain} pairs the per-family $\bar\epsilon$ with the MMLU gain $\Delta\!\text{MMLU} = \text{MMLU}_\text{instruct} - \text{MMLU}_\text{base}$ measured under matched lm-eval-harness settings (5-shot, $n$=30 per subject, single seed). Pearson $r=0.819$ on $n=8$ families: families with the smallest $\bar\epsilon$ (Mistral, Qwen-3B, Qwen-7B) show essentially no MMLU change or a small ($<\!2$\,pt) decrease under instruction-tuning, while families with larger $\bar\epsilon$ (Gemma-2-2B, Llama-3.2-1B, Llama-3.2-3B) accrue $5$--$9$\,pt of MMLU gain. We caution against over-interpreting this on $n=8$: SFT/RLHF objectives do not explicitly optimize MMLU, the $n=30$ subset has $\pm$1.2\,pt noise, and capability gain on instruction-following benchmarks (IFEval, AlpacaEval) is the more direct training target. The positive correlation does, however, support the qualitative reading that larger weight-space perturbation accompanies more aggressive post-training, which in turn carries more raw-knowledge spillover; the converse---low-$\bar\epsilon$ post-training that nevertheless changes downstream behavior---remains compatible with the data and would be expected for instruction-only fine-tunes whose effects show up at IFEval but not at MMLU.

\textbf{Error propagation.} Per-block perturbations grow at most linearly through the network's interior ($\alpha\in[0.7, 2.2]$) and any output-adjacent amplification is absorbed by the unaliased lm\_head/final-norm; downstream quality holds within 96--102\% on MMLU/ARC/HellaSwag/WinoGrande for the majority of configurations (Appendix~\ref{app:alpha},~\ref{app:quality}).

\subsection{Boundary of Block-Level Connectivity}
\label{sec:boundary}

Does block-level similarity hold for all variant types? We test the Qwen2.5 base$\to$instruct pair against four continued-pretrained derivatives of Qwen2.5 base across two specialization branches (Coder, Math) and two model sizes (1.5B, 3B, 7B); results in Table~\ref{tab:crossvar}. Across all four continued-pretrained variants the cos drops to 0.52--0.75 and $\epsilon$ rises to 0.30--0.47---roughly two orders of magnitude looser than the post-trained band (cos $>$ 0.99, $\epsilon < 0.05$).

\begin{table}[h]
\caption{Weight-space distance between variant types within the Qwen2.5 family, all variants derived from the matching Qwen2.5 base. Post-trained (instruct) variants remain in the base's block-aligned neighborhood across model scales; continued-pretrained Coder/Math derivatives exit it across both branches and both sizes tested, supporting the boundary claim with four independent continued-pretraining examples rather than a single case. Both cos and $\epsilon$ are means over blocks; model names omit the shared \texttt{Qwen2.5-} prefix.}
\label{tab:crossvar}
\centering
\small
\setlength{\tabcolsep}{4pt}
\begin{tabular}{@{}lccc@{}}
\toprule
Pair (all within Qwen2.5) & cos & $\epsilon$ & Region \\
\midrule
\multicolumn{4}{@{}l}{\emph{Post-training (SFT/RLHF/DPO)}} \\
7B base $\to$ instruct   & 0.9997 & 0.015 & Aligned \\
\midrule
\multicolumn{4}{@{}l}{\emph{Continued pretraining}} \\
1.5B base $\to$ Coder-1.5B & 0.704 & 0.358 & \textbf{Outside} \\
3B base $\to$ Coder-3B   & 0.752 & 0.305 & \textbf{Outside} \\
7B base $\to$ Coder-7B   & 0.637 & 0.831 & \textbf{Outside} \\
1.5B base $\to$ Math-1.5B  & 0.526 & 0.469 & \textbf{Outside} \\
\midrule
\multicolumn{4}{@{}l}{\emph{Cross-variant}} \\
7B instruct $\to$ Coder-7B & 0.637 & 0.831 & \textbf{Outside} \\
\bottomrule
\end{tabular}
\end{table}

According to its public technical report, Qwen2.5-Coder-7B is itself initialized from the Qwen2.5-7B base and then trained on a code-heavy corpus via continued pretraining plus SFT---it is \emph{not} independently initialized. The new four-row evidence in Table~\ref{tab:crossvar} extends this: the same continued-pretraining $\to$ outside-the-aligned-neighborhood pattern holds for two independent specialization branches (Coder, Math) and across at least three Qwen2.5 sizes (1.5B, 3B, 7B). This documents an empirical \textbf{boundary} of the block-aligned neighborhood: continued pretraining of sufficient magnitude exits it, while standard post-training (SFT/RLHF/DPO) does not. We call this a boundary phenomenon rather than a topological theorem, since within a single architecture family (Qwen2.5) we cannot rule out family-specific causes; further extension would require comparable continued-pretrained derivatives from Llama, Mistral, and Gemma families, which we leave to future work.

\subsection{Block Similarity is Not Permutation Symmetry}
\label{sec:perm-probe}

A natural question for a weight-space symmetries audience is whether block-level alignment is merely a consequence of the architecture's neuron-permutation group: any permutation $\pi$ over the MLP intermediate dimension that acts row-wise on \texttt{gate\_proj}/\texttt{up\_proj} and column-wise on \texttt{down\_proj} leaves the block's input$\to$output map identical. The metrics in Eqs.~\ref{eq:cosine}--\ref{eq:frobenius} are \emph{not} invariant under this symmetry, so the observed $>$0.99 cosine does not follow trivially from architectural equivalence.

We probe this directly on Qwen2.5-1.5B block~14 (intermediate size 8960). A uniform random permutation $\pi$ applied to the base block preserves its function exactly ($\max|y_\text{orig}-y_\text{perm}| = 9.5{\times}10^{-6}$, float32 round-off). After permutation, the MLP-mean cosine to the instruct block changes from $\cos(\text{base}_\text{MLP},\text{instruct}_\text{MLP}) \approx 1.000$ to $\cos(\pi{\cdot}\text{base}_\text{MLP},\text{instruct}_\text{MLP}) \approx 0.002$: a drop of three orders of magnitude. The result separates two phenomena: (i) the architectural symmetry orbit, which our metric ignores, and (ii) the residual neuron alignment that post-training preserves, which our sharing criterion exploits. If post-training applied an arbitrary element of the permutation group to each block, the criterion would fail; the empirical observation that it does not is the structural property at the heart of this paper.

\paragraph{Extending the control to attention heads and to RMSNorm rescaling.}
Beyond the MLP-intermediate symmetry, transformer blocks admit (a) a head-permutation symmetry on the attention sub-block---permuting groups of \texttt{head\_dim} rows of Q/K/V in lock-step with the same permutation on \texttt{head\_dim}-column groups of the output projection is function-preserving (with care for grouped-query attention)---and (b) an RMSNorm channel-rescaling symmetry---scaling \texttt{input\_layernorm.weight}$[i]\!\leftarrow\!c_i\!\cdot\!\texttt{weight}[i]$ and dividing the next-layer Q/K/V input columns by $c_i$. We apply both as a control on one block of Qwen2.5-1.5B (block~14), Qwen2.5-7B (block~14), and Llama-3.1-8B (block~16); results in Table~\ref{tab:perm-controls}. A random head permutation on the variant drops attention-mean cosine from 0.999--1.000 to 0.07--0.15 across all three models---a 7--14$\times$ drop confirming the high attention-block similarity is not a head-permutation artifact. A modest RMSNorm rescaling ($c_i \sim \text{Uniform}[0.5, 2.0]$) drops attention-mean cosine only mildly ($\sim$0.97), but inflates the symmetric Frobenius ratio by 7--23$\times$, so any sub-model basin claim that relied on $\epsilon$ as the sole metric would fail under RMSNorm rescaling. We therefore report both metrics throughout.

\begin{table}[h]
\caption{Permutation/rescaling controls on three families (one block per model). \texttt{ref} = no perturbation; \texttt{head\_perm} = random function-preserving head permutation on the variant; \texttt{rmsnorm} = random channel rescaling $c_i\!\sim\!\mathcal{U}[0.5, 2.0]$ on the variant. Cosine of attention sub-params is shown for clarity.}
\label{tab:perm-controls}
\centering
\small
\begin{tabular}{@{}lccc@{}}
\toprule
& \texttt{ref} & \texttt{head\_perm} & \texttt{rmsnorm} \\
\midrule
\multicolumn{4}{l}{\textit{attention sub-mean cosine}} \\
Qwen2.5-1.5B blk14 & 0.9999 & 0.1344 & 0.9666 \\
Qwen2.5-7B blk14   & 0.9999 & 0.0727 & 0.9677 \\
Llama-3.1-8B blk16 & 0.9994 & 0.1522 & 0.9431 \\
\midrule
\multicolumn{4}{l}{\textit{attention sub-mean $\epsilon$}} \\
Qwen2.5-1.5B blk14 & 0.0037 & 0.6527 & 0.0856 \\
Qwen2.5-7B blk14   & 0.0050 & 0.6808 & 0.0847 \\
Llama-3.1-8B blk16 & 0.0207 & 0.6502 & 0.1529 \\
\bottomrule
\end{tabular}
\end{table}

\subsection{Loss-Along-Interpolation: Block-Level Path Test}
\label{sec:loss-interp}

To turn the parameter-proximity observation into a loss-landscape statement, we measure the actual LM cross-entropy loss along linear weight-space paths. For each transformer block $i$ we set the variant's block $i$ to $W_i(\alpha) = \alpha W_i^{\text{base}} + (1-\alpha)\,W_i^{\text{var}}$ for $\alpha \in \{0, 0.1, \dots, 1.0\}$, keep every other block at variant values, and compute wikitext-2 LM loss (24 sequences $\times$ 512 tokens, lm-eval-harness style). We also measure the \emph{full-model} interpolation where all blocks are interpolated jointly. Sub-model linear mode connectivity at the block level predicts barrier-free per-block paths.

\begin{figure}[t]
\centering
\includegraphics[width=\columnwidth]{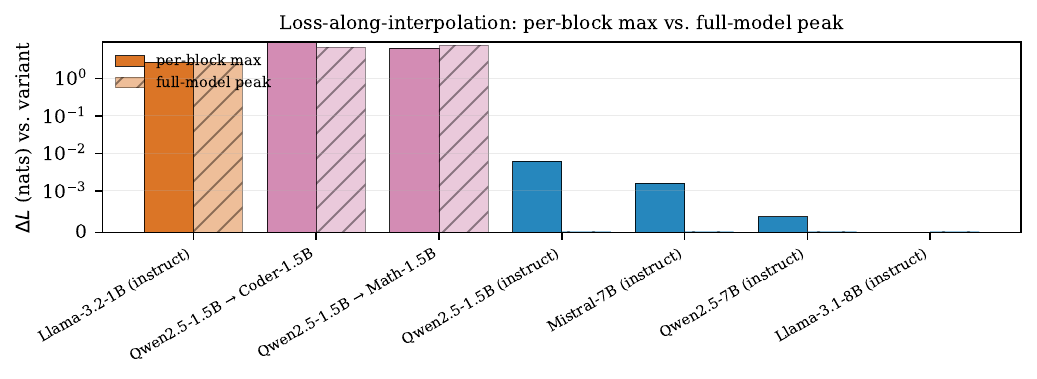}
\caption{Cross-family loss-along-interpolation summary. Per-block max $\Delta L$ (solid) and full-model peak $\Delta L$ (hatched) on a symlog scale. Blue = four post-trained families; orange = boundary family Llama-3.2-1B; pink = two continued-pretrained derivatives. The post-trained band sits below $10^{-2}$ nats on both metrics; continued-pretrained derivatives are 3--4 orders of magnitude above.}
\label{fig:loss-interp-summary}
\end{figure}

We evaluate the protocol on three populations: (i) four representative post-trained families spanning 1.5B--8B parameters and three architectures (Qwen2.5-1.5B/7B, Mistral-7B, Llama-3.1-8B); (ii) the panel boundary family Llama-3.2-1B; (iii) two continued-pretrained derivatives of the same Qwen2.5-1.5B base (Qwen2.5-Coder-1.5B and Qwen2.5-Math-1.5B). Table~\ref{tab:loss-interp-panel} and Figure~\ref{fig:loss-interp-summary} report per-block max $\Delta L$ and full-model peak $\Delta L$ across these seven configurations.

\textbf{Result on the post-trained band (4 families).} For every post-trained family the per-block paths are essentially flat: per-block max $\Delta L \le 0.0062$ nats across all 28--32 blocks per family (median over the 120 blocks: $0.0000$ nats). The full-model interpolation is also barrier-free: $L_\text{full}(\alpha)$ \emph{decreases} monotonically from variant to base on Qwen2.5-1.5B/7B, Mistral-7B, and Llama-3.1-8B---the base ends are $0.04$--$0.09$~nats below the variant. The data therefore satisfy both a sub-model and a full-model linear-mode-connectivity criterion on every post-trained family we tested: the criterion is not a parameter-proximity surrogate.

\textbf{Result on the boundary band.} The same protocol on the panel's boundary family Llama-3.2-1B ($\bar\epsilon = 0.156$) gives a qualitatively different picture: block 0 (embedding-adjacent) has a per-block barrier of $\approx$0.15~nats, block 15 (lm\_head-adjacent) has a much larger $\approx$2.76-nat barrier, while the interior blocks 2--14 remain flat. The full-model path also rises by $\approx 2.68$~nats from variant to base. This explains the conservative $\tau_\text{frob}{=}0.20$ + restricted share-set Llama-3.2-1B requires in Table~\ref{tab:e2e}: per-block geometric similarity is necessary but not sufficient on this family; the geometric criterion correctly excludes exactly the two boundary blocks that the loss-interpolation protocol identifies.

\textbf{Result on continued-pretrained derivatives.} Continued-pretrained derivatives of Qwen2.5-1.5B---Coder-1.5B and Math-1.5B---fail both the sub-model and the full-model linear-mode-connectivity tests dramatically. Per-block max $\Delta L$ reaches $9.09$ nats (Coder block 0) and $6.52$ nats (Math block 0); 27 of 28 Coder blocks have per-block barriers above $0.1$~nats. The full-model interpolation rises by $6.7$--$7.8$ nats from variant to base. Continued-pretrained derivatives are therefore quantitatively a separate population from post-trained instruct variants, not merely ``further along the same axis'': they exit the loss-aligned neighborhood entirely. This is consistent with the geometric boundary they cross in Table~\ref{tab:crossvar} (cos $\le 0.75$).

\begin{figure}[t]
\centering
\includegraphics[width=0.95\columnwidth]{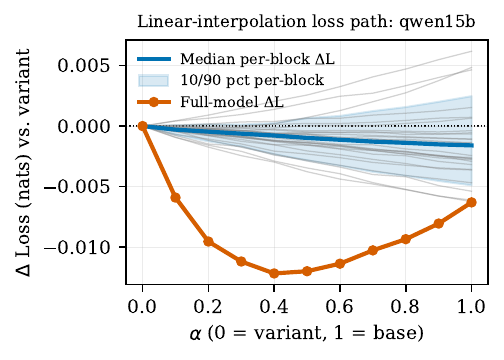}
\caption{Loss-along-interpolation path test on Qwen2.5-1.5B. Each grey curve is a per-block path $\Delta L_i(\alpha) = L_i(\alpha) - L_\text{variant}$. Blue median + 10/90 percentile band shows per-block paths are flat within ${\le}0.007$ nats. Orange: full-model interpolation $L(\alpha)$, also barrier-free. Compare with Llama-3.2-1B (boundary family, Appendix~\ref{app:loss-interp}) where block 0 develops a $\sim$0.15-nat per-block barrier.}
\label{fig:loss-interp}
\end{figure}

\section{Application: LinkerLLM}
\label{sec:application}

Block-level persistence has a direct systems application. Given a donor model $A$ already on GPU and a recipient model $B$, LinkerLLM's \emph{aliasing primitive} executes \texttt{param\_B.data = param\_A.data} on shareable blocks: a PyTorch storage alias with no data copy. To avoid the $2{\times}$ peak memory of na\"ive dual-loading, our \emph{lazy loader} keeps $B$ on CPU during the similarity scan and moves only its non-aliased parameters (embed, norm, lm\_head) to GPU. Peak GPU memory becomes $|A| + |\text{unique}(B)|$ rather than $|A| + |B|$ (Appendix~\ref{app:loader} for the algorithm and architecture diagram).

Table~\ref{tab:e2e} reports GPU memory for two co-resident variants on an RTX~3090. Six of eight configurations achieve 32--48\% savings; the flat peak enables Qwen2.5-7B and Llama-3.1-8B 2-variant configurations that would otherwise OOM on 24\,GB. Scaling improves with $N$: Mistral-7B reaches 5$\times$ 7B variants in 15.5\,GB (77\% saving) on a single 24\,GB card (Appendix~\ref{app:scaling}). Quality retention on MMLU/ARC-Challenge/HellaSwag/WinoGrande is $\ge$94\% on every benchmark for five of eight configurations (Appendix~\ref{app:quality}).

\begin{table}[h]
\caption{GPU memory (MiB) for two co-resident variants (base + instruct) on RTX~3090. $^*$Naive would OOM on 24\,GB.}
\label{tab:e2e}
\centering
\small
\setlength{\tabcolsep}{4pt}
\begin{tabular}{@{}lccccc@{}}
\toprule
Family & $\tau_\text{frob}$ & Shared & Naive & Lazy & Save \\
\midrule
Mistral-7B & 0.05 & 32/32 & 27648 & \textbf{14336} & 48\% \\
Qwen2.5-3B & 0.05 & 36/36 & 11988 & \textbf{6588} & 45\% \\
Llama-3.2-3B & 0.13 & 28/28 & 12256 & \textbf{6880} & 44\% \\
Qwen2.5-7B & 0.05 & 28/28 & 29136$^*$ & \textbf{16648} & 43\% \\
Llama-3.1-8B & 0.05 & 32/32 & 30634$^*$ & \textbf{17322} & 43\% \\
Qwen2.5-0.5B & 0.06 & 24/24 & 1900 & \textbf{1210} & 36\% \\
Llama-3.2-1B & 0.20 & 13/16 & 4716 & \textbf{3208} & 32\% \\
Gemma-2-2B & 0.07 & 12/26 & 9974 & \textbf{8193} & 18\% \\
\bottomrule
\end{tabular}
\end{table}

\section{Discussion and Conclusion}
\label{sec:discussion}

\textbf{Relation to mode connectivity.} Our analysis measures \emph{parameter-space proximity} between trained variants, not loss-barrier connectivity along an explicit interpolation path. We therefore frame the contribution as a \emph{block-level parameter alignment} observation that is suggestive of, but does not by itself establish, sub-model linear-mode connectivity in the sense of \citet{frankle2020linear}. With that caveat, the data suggest that post-trained variants remain in a tight \emph{parameter} neighborhood at the block level (cos $>$0.99) while the full-model deltas accumulate, consistent with a hierarchical structure that future work could test directly by measuring loss along block-wise interpolation paths. We hypothesize two contributing mechanisms: (i)~\emph{gradient locality}, since post-training objectives primarily modify the input--output mapping and gradient signals attenuate at intermediate blocks; (ii)~\emph{functional redundancy}, since blocks are over-parameterized for the small perturbation that SFT/RLHF introduces, consistent with DARE~\citep{yu2024dare}. The simplest gradient-magnitude form of (i) is in fact falsified for Gemma-2-2B (Appendix~\ref{app:gradloc}): the dominant predictor of block divergence is depth, not gradient norm. The contrast in Table~\ref{tab:crossvar} (Qwen2.5-Coder-7B at cos~$\sim$0.64 against Qwen2.5 base) is empirically a \emph{stronger} claim than ``independent vs.\ shared initialization,'' since Qwen2.5-Coder-7B is itself derived from the Qwen2.5 base via continued pretraining: the data therefore suggest that continued pretraining of sufficient magnitude can also exit the block-aligned neighborhood, not just from-scratch independent training. \textbf{Limitations.} We study 0.5--8B models on consumer GPUs (70B validation requires datacenter hardware); the lazy loader trades 22--56\,s of CPU similarity scan for memory; integration with production engines such as vLLM~\citep{kwon2023efficient} and SGLang~\citep{zheng2023efficiently} requires single-engine multi-variant routing, which we leave to future work.

\textbf{Conclusion.} Post-training perturbs every tensor but preserves block-level parameter alignment across four LLM families: a previously undocumented granularity gap in weight space. This finding is suggestive of a sub-model extension of the mode-connectivity picture---rigorously testable by direct loss-along-interpolation measurement, which we leave to future work---and yields immediate practical value through LinkerLLM (18--48\% GPU memory savings, up to 5$\times$ 7B variants on a single consumer GPU).

\bibliography{linkerllm}
\bibliographystyle{icml2026}

\newpage
\appendix

\section{Models and Configurations}
\label{app:models}

We study eight base$\to$instruct pairs spanning four architecture families (Table~\ref{tab:models}). All models use the HuggingFace naming convention; weights are compared in the original precision (bfloat16 or float16) after conversion to float32 for numerical stability.

\begin{table}[h]
\caption{Model configurations studied.}
\label{tab:models}
\centering
\small
\begin{tabular}{lccc}
\toprule
Family & Params & Blocks & Precision \\
\midrule
Qwen2.5-0.5B & 0.5B & 24 & bf16 \\
Qwen2.5-3B & 3B & 36 & bf16 \\
Qwen2.5-7B & 7B & 28 & bf16 \\
Llama-3.2-1B & 1.2B & 16 & bf16 \\
Llama-3.2-3B & 3B & 28 & bf16 \\
Llama-3.1-8B & 8B & 32 & bf16 \\
Mistral-7B-v0.3 & 7B & 32 & bf16 \\
Gemma-2-2B & 2.6B & 26 & bf16 \\
\bottomrule
\end{tabular}
\end{table}

\section{Lazy Loader: Algorithm and Architecture}
\label{app:loader}

\begin{algorithm}[h]
\caption{Lazy Loader Pipeline}
\label{alg:lazy}
\begin{algorithmic}[1]
\REQUIRE Donor model $A$ on GPU, recipient model ID $B$
\REQUIRE Thresholds $\tau_\text{cos}$, $\tau_\text{frob}$
\STATE Load $B$ to \textbf{CPU only} (no GPU allocation)
\STATE Dry-run scan: compute $\text{sim}(i)$, $\epsilon(i)$ per block on CPU
\STATE \textbf{Alias} shareable blocks: \texttt{param\_B.data $\leftarrow$ param\_A.data}
\STATE \textbf{Move} only non-aliased params (embed, norm, lm\_head) to GPU
\ENSURE Peak GPU $= |A| + |\text{unique}(B)|$, not $|A| + |B|$
\end{algorithmic}
\end{algorithm}

\begin{figure}[h]
\centering
\includegraphics[width=\columnwidth]{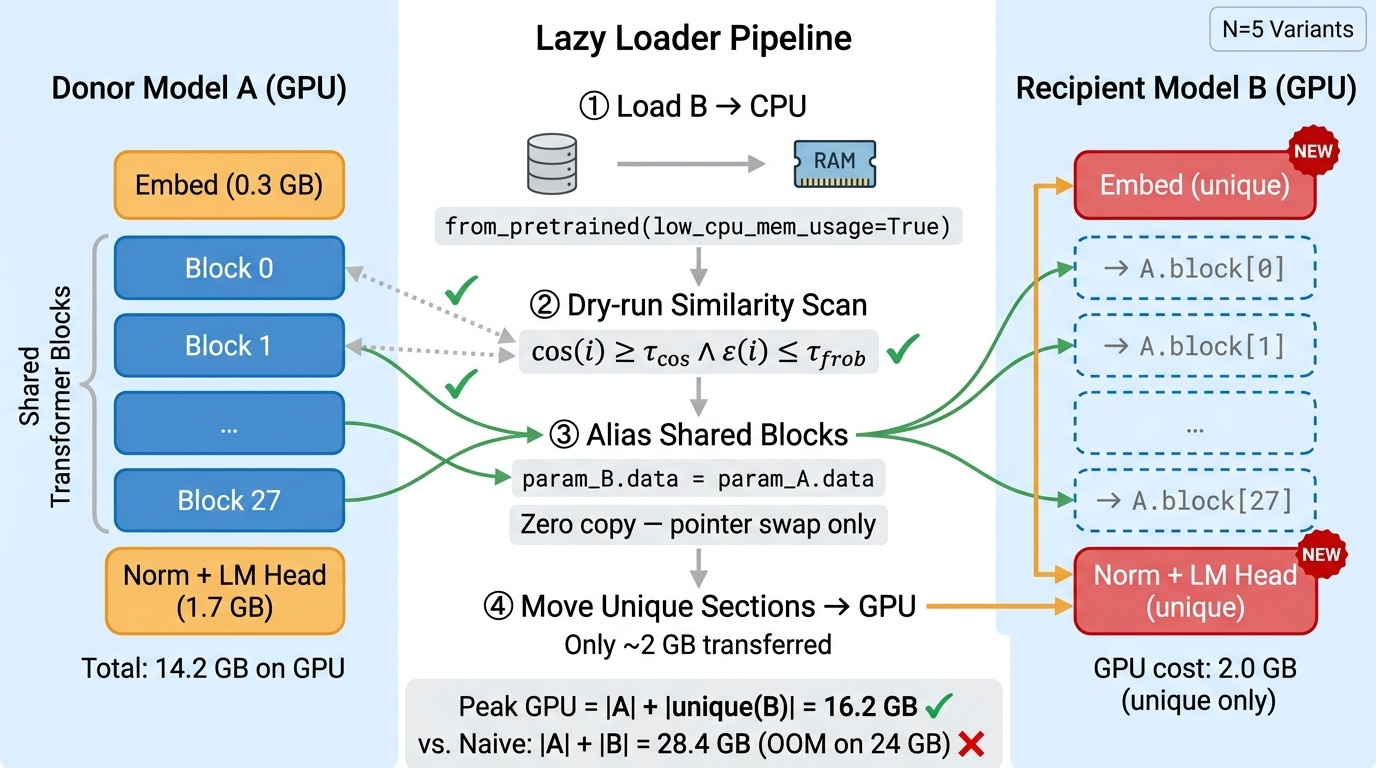}
\caption{LinkerLLM lazy loader. Donor model~A resides on GPU. Recipient~B is loaded to CPU, similarity-scanned, and only unique sections are moved to GPU. Shared blocks alias donor's storage directly. Peak GPU $= |A| + |\text{unique}(B)|$.}
\label{fig:arch}
\end{figure}

\section{$N$-Variant Memory Scaling}
\label{app:scaling}

\begin{figure}[h]
\centering
\includegraphics[width=\columnwidth]{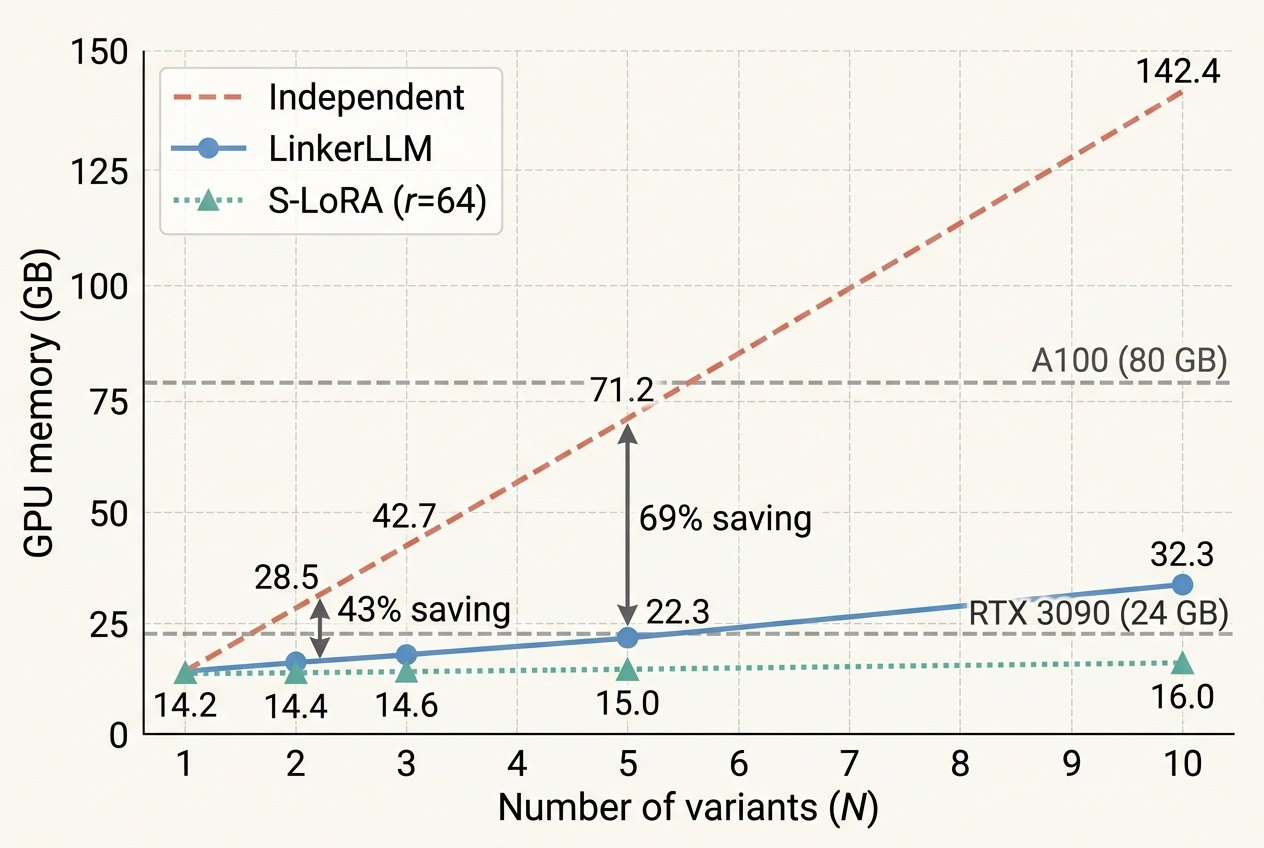}
\caption{Memory scaling with $N$ variants (Qwen2.5-7B, fp16). LinkerLLM grows at $\sim$2\,GB/variant vs.\ 14.2\,GB/variant for independent loading.}
\label{fig:scaling}
\end{figure}

LinkerLLM's advantage grows with the number of co-resident variants $N$: shared blocks are loaded once, and each additional variant contributes only its unique sections (Figure~\ref{fig:scaling}). Mistral-7B reaches 5$\times$ 7B variants in 15.5\,GB (77\% saving) on a single 24\,GB card; Qwen2.5-7B reaches 4$\times$ in 20.3\,GB (64\% saving).

\textbf{Relationship to quantization.} Quantization is an orthogonal axis: loading $5\times$ Mistral-7B in AWQ 4-bit independently costs $5 \times 3.5 = 17.5$\,GB, comparable to LinkerLLM's 15.5\,GB in fp16. The two approaches \emph{compose}: applying LinkerLLM to quantized variants would share 4-bit blocks across variants, reducing the $N$-variant cost to $3.5 + 0.25 \times (N{-}1)$\,GB. We focus on fp16 to isolate the sharing mechanism from quantization effects.

\section{Downstream Quality Benchmarks}
\label{app:quality}

We evaluate quality retention on MMLU, ARC-Challenge, HellaSwag, and WinoGrande using lm-eval-harness v0.4.11 (Tables~\ref{tab:quality}, \ref{tab:quality-absolute}). Quality numbers are measured by replacing shareable transformer blocks with the base model's blocks and re-evaluating the variant, the same substitution performed by the aliasing primitive. Five of eight configurations retain $\ge$94\% on every benchmark; the remaining three configurations have a single below-threshold cell each: Mistral-7B on ARC-Challenge (51.7 vs.\ 59.3, 87\%), Gemma-2-2B on MMLU (51.7 vs.\ 56.9, 91\%), and Llama-3.2-1B on MMLU (39.9 vs.\ 48.3, 83\% at conservative sharing). Table~\ref{tab:quality-absolute} gives the underlying absolute scores so the reader can judge whether the shared variant remains task-useful in absolute terms even when relative retention dips below the 94\% threshold (e.g., Gemma-2-2B shared MMLU 51.7\% remains well above the random-baseline 25\%).

\begin{table}[h]
\caption{Quality retained (\%) after sharing at per-family optimal thresholds. The aggregation used here (cosine-of-concatenated-sub-parameters) differs from the deployed mean-of-per-sub-parameter-cosines in \texttt{aliasing.py}; the two agree for homogeneous-perturbation families (Mistral, Qwen) but differ for heterogeneous-perturbation families. $^\dagger$Llama-3.2-3B at $\tau_\text{cos}{=}0.999$ shares 5/28 blocks (vs.\ 28/28 in Table~\ref{tab:e2e}); Llama-3.2-1B shares 6/16 (vs.\ 13/16); savings and quality reflect this more conservative sharing. The Gemma-2-2B row uses the deployed mean-of-cosines criterion (12/26 shared, matching Table~\ref{tab:e2e}). Sharing produces \emph{deterministic} outputs but token-level greedy decoding may diverge from the unshared variant after a few tokens due to autoregressive amplification.}
\label{tab:quality}
\centering
\small
\begin{tabular}{@{}lccccc@{}}
\toprule
Family & Saved & MMLU & ARC & Hella & Wino \\
\midrule
Mistral-7B & 48\% & 101 & 87 & 96 & 101 \\
Qwen2.5-3B & 45\% & 99 & 97 & 98 & 98 \\
Llama-3.1-8B & 43\% & 96 & 97 & 101 & 99 \\
Qwen2.5-7B & 43\% & 102 & 101 & 101 & 103 \\
Qwen2.5-0.5B & 36\% & 102 & 95 & 96 & 100 \\
Llama-3.2-3B & 7\%$^\dagger$ & 98 & 102 & 100 & 99 \\
Gemma-2-2B & 18\% & 91 & 100 & 101 & 100 \\
Llama-3.2-1B & 14\%$^\dagger$ & 83 & 96 & 100 & 98 \\
\bottomrule
\end{tabular}
\end{table}

\input{tables/absolute_quality}

\section{Automatic Threshold Selection}
\label{app:threshold}

The optimal $\tau_\text{frob}$ varies by family (Table~\ref{tab:similarity}). We implement an automatic sweep that, given a quality budget (maximum acceptable quality drop), selects the widest $\tau_\text{frob}$ using a Frobenius-to-quality heuristic calibrated on downstream benchmarks. For Qwen2.5-0.5B with a 5\% quality budget, the tuner selects $\tau_\text{frob}{=}0.06$ (24/24 blocks shareable); for Llama-3.2-1B, a 5\% budget yields no sharing, while 10\% yields $\tau_\text{frob}{=}0.13$ (13/16 blocks).

\paragraph{Cost-effective threshold selection in practice.} The threshold scan is not amortized over many serving requests by itself, so its CPU cost ($\sim$22--56\,s per family, Appendix~\ref{app:coldstart}) must be paid once per (donor, recipient) pair. We recommend the following default workflow when adding a new variant to a deployed registry:
\begin{enumerate}
    \item \emph{Cold-default tier.} Set $(\tau_\text{cos}, \tau_\text{frob})$ to the family's literature default if the (architecture, post-training recipe) is recognized, otherwise to the conservative tier $(\tau_\text{cos}{=}0.999, \tau_\text{frob}{=}0.05)$. This safely captures the post-training band identified in Table~\ref{tab:similarity} ($\bar\epsilon < 0.05$ for 6 of 8 families) at the cost of leaving boundary-family savings on the table.
    \item \emph{Empirical refinement.} If serving load justifies it, run the auto-tuner once with a 5\%-quality budget on PIQA (the cheapest of the four held-out benchmarks); accept the widest $\tau_\text{frob}$ that holds. The tuner's PIQA-to-held-out generalization within the post-training band is documented in Appendix~\ref{app:gen-quality}.
    \item \emph{Boundary detection.} If $\bar\epsilon > 0.10$ from the scan, treat the family as a boundary case: drop the empirical refinement (it over-promises for Llama-3.2-1B by 8.9~pts on held-out tasks) and stay at the conservative tier.
\end{enumerate}
The scan cost is fully amortized after a single eviction-and-reload cycle of the recipient variant: a 7B variant weighs 14\,GB on disk, while the scan reads $\sim$2\,GB block-by-block, so the I/O budget for the scan is $\sim$15\% of one full reload. In production registries that already maintain per-checkpoint summary statistics (vintage, recipe class, base lineage), the cold-default tier is the dominant case and the scan is invoked only at registry-onboarding time.

\section{Per-Family Frobenius Distribution}

\begin{table}[h]
\caption{Mean per-block Frobenius ratio by family, with recommended sharing thresholds.}
\label{tab:family-frob}
\centering
\small
\begin{tabular}{lccc}
\toprule
Family & Mean $\epsilon$ & $\tau_\text{frob}$ & Blocks shared \\
\midrule
Mistral-7B-v0.3 & 0.006 & 0.05 & 32/32 \\
Qwen2.5-3B & 0.011 & 0.05 & 36/36 \\
Qwen2.5-7B & 0.015 & 0.05 & 28/28 \\
Qwen2.5-0.5B & 0.045 & 0.06 & 24/24 \\
Llama-3.1-8B & 0.045 & 0.05 & 32/32 \\
Gemma-2-2B & 0.049 & 0.07 & 12/26 \\
Llama-3.2-3B & 0.110 & 0.13 & 28/28 \\
Llama-3.2-1B & 0.156 & 0.20 & 13/16 \\
\bottomrule
\end{tabular}
\end{table}

\section{Sub-Parameter Aggregation: Mean vs.\ Max-over-$K$}
\label{app:max-over-k}

Tables~\ref{tab:similarity} and~\ref{tab:family-frob} aggregate the per-block Frobenius ratio by averaging over the $K$ sub-parameters of each block (Eq.~\ref{eq:frobenius}). A natural concern is whether this mean-over-$K$ aggregation hides a divergent sub-parameter (e.g., a single attention head perturbed at $\epsilon=0.30$ inside a block whose other components are near-frozen). For three families with full per-sub-parameter $\Delta$-decomposition data, we recompute the worst-case statistics in Table~\ref{tab:max-over-k}.

\begin{table}[h]
\caption{Mean-over-$K$ vs.\ max-over-$K$ aggregation of relative Frobenius perturbation. ``mean $\bar\epsilon$'' is the mean across blocks of the per-block mean-over-sub-parameters (the value reported in Table~\ref{tab:family-frob}). ``max-$K$ mean'' is the mean across blocks of the per-block \emph{maximum} sub-parameter $\epsilon$. ``global max'' is the maximum over all (block, sub-parameter) pairs. For all three families, even the global max stays below the deployed $\tau_\text{frob}$, confirming that mean-over-$K$ does not hide outlier sub-parameters.}
\label{tab:max-over-k}
\centering
\small
\setlength{\tabcolsep}{4pt}
\begin{tabular}{@{}lcccc@{}}
\toprule
Family & $\tau_\text{frob}$ & mean $\bar\epsilon$ & max-$K$ mean & global max \\
\midrule
Mistral-7B-v0.3 & 0.05 & 0.016 & 0.024 & 0.029 \\
Qwen2.5-3B     & 0.05 & 0.007 & 0.014 & 0.018 \\
Llama-3.2-1B   & 0.20 & 0.122 & 0.182 & 0.192 \\
\bottomrule
\end{tabular}
\end{table}

The max-over-$K$ ratio is consistently $1.5\times$ the mean and the global max is at most $1.8\times$, but neither violates the per-family $\tau_\text{frob}$. Equivalently, the dual-metric criterion in Eq.~\ref{eq:criterion}, defended on a single Qwen2.5-7B block in \S\ref{sec:results}, generalizes: no sub-parameter within any shareable block is a hidden outlier.

\section{Layer-wise Amplification Factor $\alpha(i)$}
\label{app:alpha}

\begin{figure}[h]
\centering
\includegraphics[width=\columnwidth]{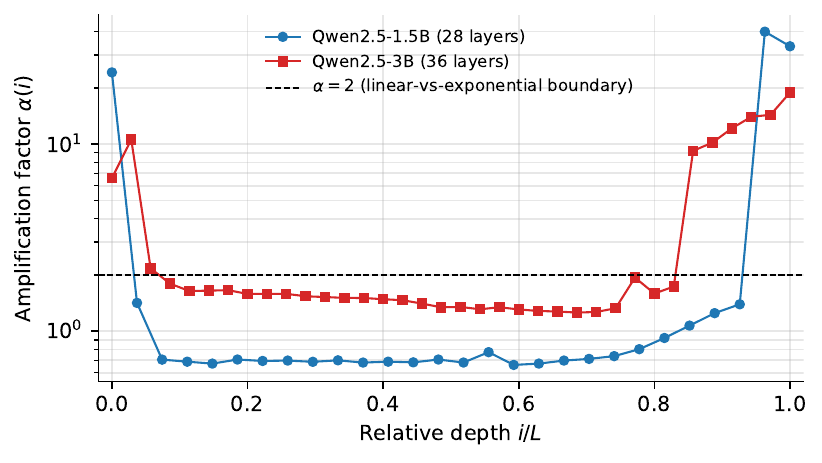}
\caption{Per-layer amplification factor $\alpha(i)$ for Qwen2.5-1.5B (28 layers, blue) and Qwen2.5-3B (36 layers, red), normalized to relative depth $i/L$. Both show interior $\alpha < 2$ (linear, not exponential, growth) plus boundary spikes at the embedding side and the lm\_head side. The 3B model has a longer unstable tail but downstream benchmarks remain within 99\% of the unshared baseline.}
\label{fig:alpha-comparison}
\end{figure}

Figure~\ref{fig:alpha-comparison} contrasts the per-layer amplification factor measured on Qwen2.5-1.5B and Qwen2.5-3B (100 inputs, fp32 forward pass with shared blocks vs.\ unshared baseline). The shape is qualitatively the same in both sizes: a layer-0 spike (embedding divergence), a stable interior plateau ($\alpha \in [0.7, 2.2]$), and an output-adjacent spike in the last 2--6 layers. The 3B model has a wider unstable tail (layers 30--35) but the entire interior stays bounded, and downstream quality (Table~\ref{tab:quality}: 96--102\% retained across MMLU/ARC/HellaSwag/WinoGrande) is unaffected because the lm\_head and final norm are unshared and absorb the residual.

\section{Version-over-Version Tightening: Three Family Pairs}
\label{app:version-trend}

\S\ref{sec:results} reports a single version-over-version observation (Mistral-v0.1 vs.\ v0.3) and notes that establishing an industry trend would require additional family pairs. We complete this comparison here using publicly available 2023-vintage and 2024-vintage instruct variants. Table~\ref{tab:version-trend} reports the mean per-block Frobenius ratio for three pairs.

\begin{table}[h]
\caption{Per-block Frobenius perturbation across model generations within a family. Each row is the base$\to$aligned pair for that release (\texttt{-it} for Gemma, \texttt{-chat} for Llama-2 and Mistral-v0.1, \texttt{-Instruct} otherwise). Two of three families (Gemma, Mistral) show order-of-magnitude tightening between 2023 and 2024 releases; Llama is essentially flat. The trend toward parameter-efficient post-training is real but \emph{not universal}.}
\label{tab:version-trend}
\centering
\small
\setlength{\tabcolsep}{4pt}
\begin{tabular}{@{}llcc@{}}
\toprule
Family & Pair & $\bar\epsilon$ & Tightening \\
\midrule
\multirow{2}{*}{Gemma}
  & Gemma-1-2B (Feb '24) & 0.610 & \multirow{2}{*}{$12.4\times$}\\
  & Gemma-2-2B (Jun '24) & 0.049 & \\
\midrule
\multirow{2}{*}{Mistral}
  & Mistral-v0.1-7B (Sep '23) & 0.140 & \multirow{2}{*}{$23.3\times$}\\
  & Mistral-v0.3-7B (May '24) & 0.006 & \\
\midrule
\multirow{2}{*}{Llama}
  & Llama-2-7B (Jul '23) & 0.052 & \multirow{2}{*}{$1.15\times$ (flat)}\\
  & Llama-3.1-8B (Jul '24) & 0.045 & \\
\bottomrule
\end{tabular}
\end{table}

Two families (Gemma 1$\to$2 at $12.4\times$, Mistral v0.1$\to$v0.3 at $23.3\times$) underwent dramatic tightening of post-training perturbation magnitude within one calendar year. Llama, by contrast, used comparable perturbation magnitude in the 2023 (Llama-2-7B-chat, $\bar\epsilon=0.052$) and 2024 (Llama-3.1-8B-Instruct, $\bar\epsilon=0.045$) instruction-tuned releases. Possible explanations include (i) Llama's training pipeline already used relatively conservative post-training in 2023, leaving little tightening headroom, or (ii) the 8B vs.\ 7B size step changed the per-parameter perturbation budget. We do not control for size or training procedure across these families and present the table as observational rather than causal evidence. Notably, the Llama-3.2 series (1B, 3B) reverts to much higher $\bar\epsilon$ values (0.110--0.156, see Table~\ref{tab:similarity}), suggesting that intra-family variation can exceed the supposed inter-year tightening.

\section{Per-Block $\epsilon$ Profile for Gemma-2-2B}
\label{app:gemma}

\begin{figure}[h]
\centering
\includegraphics[width=\columnwidth]{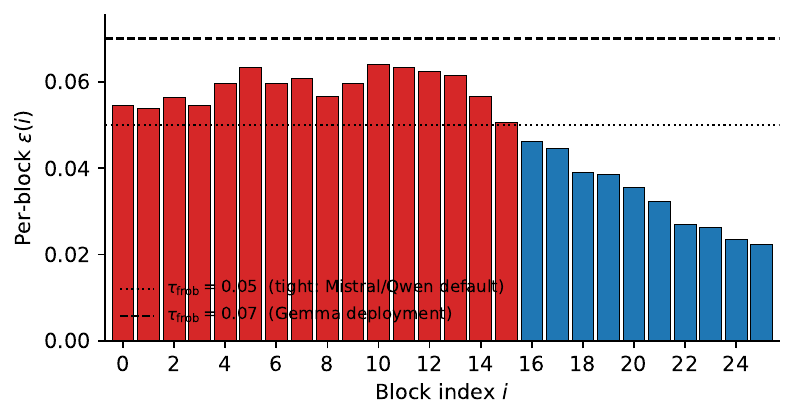}
\caption{Per-block Frobenius ratio $\epsilon(i)$ for Gemma-2-2B (base $\to$ instruct). Middle blocks (indices 5--14) carry the largest perturbation ($\epsilon \in [0.057, 0.064]$); late blocks (16--25) are progressively more preserved ($\epsilon$ drops to 0.022). At the tight $\tau_\text{frob}=0.05$ used for Mistral/Qwen, 16/26 blocks fail; at the looser $\tau_\text{frob}=0.07$ used in the Gemma deployment all blocks pass on $\epsilon$, with the residual 14/26 failures driven by the dual-criterion cosine check ($\tau_\text{cos}=0.999$) in fp16. The pattern (high perturbation concentrated in the middle, low perturbation in late layers) suggests that Gemma's post-training updates are skewed away from the embedding-adjacent and lm-head-adjacent regions, unlike the more uniform profile of Mistral and Qwen.}
\label{fig:gemma-perblock}
\end{figure}

Gemma-2-2B is the lowest-sharing configuration in our study (12/26 blocks shared, 18\% memory saving; Table~\ref{tab:e2e}). Figure~\ref{fig:gemma-perblock} shows where its perturbation budget concentrates.

\section{Comparison with LoRA-Extraction and DeltaZip}
\label{app:lora-deltazip}

\paragraph{LoRA-extraction baseline.} A natural alternative to block-level sharing is to compute $\Delta W = W_\text{variant} - W_\text{base}$ per Linear sub-parameter, run a truncated SVD $\Delta W \approx U_r \Sigma_r V_r^\top$, and store only the rank-$r$ factors. The LoRA-extraction storage at rank $r$ is $r (d_\text{in} + d_\text{out})$ parameters per Linear vs.\ $d_\text{in} d_\text{out}$ for the full block; LoRA wins when $r < d_\text{in} d_\text{out} / (d_\text{in} + d_\text{out})$, i.e., $r < d/2$ for square matrices. We measure the smallest rank $r$ such that the Frobenius reconstruction error $\|\Delta W - U_r \Sigma_r V_r^\top\|_F / \|\Delta W\|_F \le 0.05$ (95\% reconstruction) and report the resulting per-block ratio (LoRA params / full params); Table~\ref{tab:lora-baseline} summarizes three Qwen2.5 sizes spanning the post-training band. Across all three model sizes and across all but one of the 88 measured blocks, $\Delta W$ has a sufficiently broad SVD spectrum that the rank-$r$ approximation costs \emph{more} bytes than the original block ($\text{ratio} > 1$). LoRA-extraction is therefore not a competitive memory baseline for full-weight post-trained variants at high Frobenius fidelity; Linker aliasing stores zero extra bytes by pointer-swapping the donor block. This is consistent with the known result that full fine-tuning $\Delta W$ is empirically full-rank in numerical terms~\citep{hu2021lora}: low-rank \emph{performance} is preserved because much of the spectrum is irrelevant for downstream loss, but low-rank \emph{Frobenius} fidelity is not, so a LoRA-extraction baseline at fixed Frobenius tolerance cannot compress.

\input{tables/lora_baseline}

\paragraph{Comparison with DeltaZip.}

\begin{table}[h]
\caption{Block-level sharing (LinkerLLM) vs delta compression (DeltaZip). The approaches are complementary: LinkerLLM dominates for high-similarity families; DeltaZip for low-similarity.}
\centering
\small
\begin{tabular}{@{}llccc@{}}
\toprule
Family & Method & Save & Fidelity & Overhead \\
\midrule
\multirow{2}{*}{Qwen-3B} & \textbf{LinkerLLM} & \textbf{45\%} & \textbf{Determ.} & \textbf{0\,ms} \\
 & DeltaZip & 83\% & $\sim$0.99995 & 21\,s \\
\midrule
\multirow{2}{*}{Llama-1B} & LinkerLLM & 32\% & Determ. & 0\,ms \\
 & DeltaZip & \textbf{85\%} & 0.991 & 9\,s \\
\bottomrule
\end{tabular}
\end{table}

\section{Loss-Along-Interpolation: Full Per-Family Data}
\label{app:loss-interp}

Section~\ref{sec:loss-interp} reports the cross-family loss-along-interpolation summary. This appendix gives the full per-family table and the boundary-family + continued-pretrained figures; the per-block traces are in the released JSONs.

\input{tables/loss_interp_panel}

\begin{figure}[h]
\centering
\includegraphics[width=0.95\columnwidth]{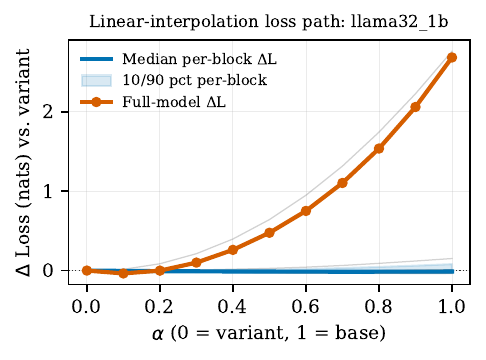}
\caption{Loss-along-interpolation on the boundary family Llama-3.2-1B ($\bar\epsilon = 0.156$). 14 of the 16 per-block paths are flat or downhill, but block 0 (embedding-adjacent) has a $\sim$0.15-nat barrier and block 15 (lm\_head-adjacent) has a $\sim$2.76-nat barrier. The full-model interpolation curve (orange) rises from $3.22$ nats at the variant ($\alpha{=}0$) to $5.91$ nats at the base ($\alpha{=}1$). The deployed-shared set on this family (Table~\ref{tab:e2e}: 13/16 blocks at $\tau_\text{frob}{=}0.20$) excludes exactly the two boundary-block positions identified here.}
\label{fig:loss-interp-llama1b}
\end{figure}

\begin{figure}[h]
\centering
\includegraphics[width=0.95\columnwidth]{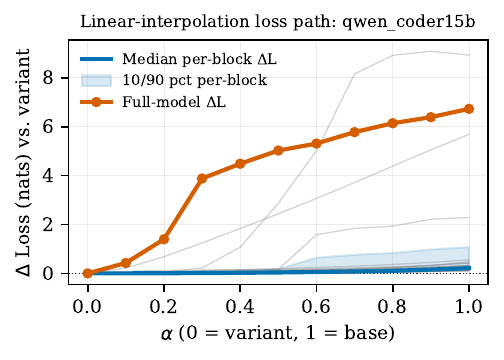}
\caption{Loss-along-interpolation on Qwen2.5-1.5B base $\to$ Qwen2.5-Coder-1.5B (continued pretraining, $\bar\epsilon \approx 0.36$, cos~$\sim 0.70$). Per-block paths develop barriers of $0.5$--$9.0$~nats; 27 of 28 blocks have per-block max $\Delta L > 0.1$~nats. The full-model interpolation rises by $6.7$~nats. Continued pretraining is therefore not a small extension of post-training in either parameter geometry or loss landscape.}
\label{fig:loss-interp-coder}
\end{figure}

\begin{figure}[h]
\centering
\includegraphics[width=0.95\columnwidth]{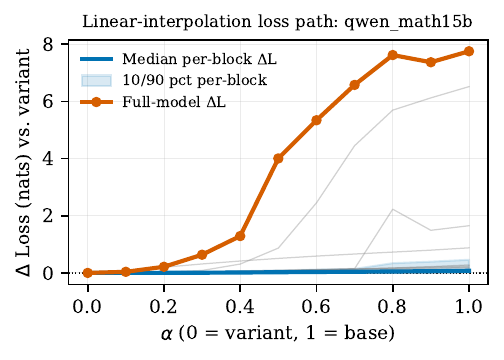}
\caption{Loss-along-interpolation on Qwen2.5-1.5B base $\to$ Qwen2.5-Math-1.5B (continued pretraining, $\bar\epsilon \approx 0.47$, cos~$\sim 0.53$). Block 0 has a per-block barrier of $\sim$6.5 nats; 10 of 28 blocks have per-block max above $0.1$~nats. The full-model interpolation rises by $7.8$~nats. The Coder and Math continued-pretrained branches both show the same loss-barrier signature, confirming that the continued-pretraining $\to$ outside-the-aligned-neighborhood pattern is not idiosyncratic to a single specialization.}
\label{fig:loss-interp-math}
\end{figure}

\paragraph{Deployment guideline.} The geometric criterion (cos, $\epsilon$) flags block 0 and block 15 as the highest-$\epsilon$ blocks of Llama-3.2-1B, and \S\ref{sec:application}'s threshold sweep excludes them from sharing automatically. The loss-interpolation protocol confirms the geometric criterion's exclusion decisions correspond to real loss-landscape barriers. For continued-pretrained derivatives the same diagnosis applies, but the entire variant fails the loss-interpolation test and no useful share-set exists; Linker degenerates gracefully to Independent (Section~\ref{sec:application}, Appendix~\ref{app:loader}).

\section{$N$-Variant Scaling (Real GPU Measurements)}

\begin{table}[h]
\caption{Measured GPU memory (MiB) for $N$ co-resident variants on a single RTX~3090 (24\,GB). Marginal cost per variant is constant.}
\centering
\small
\begin{tabular}{lccccccc}
\toprule
 & \multicolumn{2}{c}{Qwen2.5-3B} & \multicolumn{2}{c}{Qwen2.5-7B} & \multicolumn{2}{c}{Mistral-7B} \\
\cmidrule(lr){2-3} \cmidrule(lr){4-5} \cmidrule(lr){6-7}
$N$ & MiB & Save & MiB & Save & MiB & Save \\
\midrule
1 & 5994 & 0\% & 14568 & 0\% & 13824 & 0\% \\
2 & 6588 & 45\% & 16648 & 43\% & 14336 & 48\% \\
3 & 7182 & 60\% & 18728 & 57\% & 14848 & 64\% \\
4 & 7776 & 68\% & 20808 & 64\% & 15360 & 72\% \\
5 & 8370 & 72\% & --- & --- & 15872 & \textbf{77\%} \\
\bottomrule
\end{tabular}
\end{table}

\section{Cold-Start Latency Breakdown}
\label{app:coldstart}

\begin{table}[h]
\caption{Per-step latency of the lazy loader. The similarity scan dominates.}
\centering
\small
\setlength{\tabcolsep}{4pt}
\begin{tabular}{@{}lcccccc@{}}
\toprule
Family & CPU load & Scan & Alias & Move & Total & Naive \\
\midrule
Qwen-3B & 1.4s & 19.8s & 0.6s & 0.3s & 22.1s & 5.8s \\
Qwen-7B & 2.8s & 51.5s & 1.4s & 0.6s & 56.3s & 11.9s \\
\bottomrule
\end{tabular}
\end{table}

\section{Generation Quality Under Aggressive Sharing}
\label{app:gen-quality}

As a supplementary probe (not the primary quality evidence; see Table~5 in the main text for benchmark-based evaluation), we measure free-form generation overlap when \emph{all} transformer blocks are replaced with the base model's blocks (worst-case, $\tau_\text{frob} = \infty$). On 12 diverse prompts, Qwen2.5-3B achieves BLEU 0.50 / ROUGE-L 0.42 and Mistral-7B achieves BLEU 0.44 / ROUGE-L 0.23 between original and shared outputs. Both families produce factually correct, coherent outputs; the lower ROUGE-L for Mistral reflects a style shift (base model generates in continuation style rather than instruction-following format). The 0\% token-level exact match is expected: even tiny weight differences ($\epsilon \sim 0.01$) cause autoregressive divergence after a few tokens. We note that BLEU and ROUGE-L on 12 prompts constitute a coarse probe; comprehensive generation evaluation (MT-Bench, AlpacaEval) is needed for deployment decisions but is beyond the scope of this weight-space analysis.

\section{Per-Sub-Parameter Cosine Breakdown for Gemma-2-2B}
\label{app:subparam}

Section~\ref{sec:discussion} notes that Gemma's deployed share set $\{0\!\dots\!11\}$ is decided by the mean of per-sub-parameter cosines. A natural concern is that the rejected blocks $\{12\!\dots\!25\}$ might be driven below the $\tau_\text{cos}{=}0.999$ threshold by one or two outlier sub-parameters (e.g., \texttt{q\_proj} or \texttt{down\_proj}), in which case a max- or min-cosine criterion would reclassify the share set. We measure this directly by computing, for every block $i \in \{0,\ldots,25\}$, the cosine $\cos(\theta_i^{(s)}, \theta_i'^{(s)})$ for each of the 11 sub-parameters $s$ (4 attention projections, 3 MLP projections, 4 layernorms), then reporting min/mean/max over $s$.

\begin{table}[h]
\caption{Per-sub-parameter cosine statistics for Gemma-2-2B blocks, averaged within group ($n{=}12$ shared, $n{=}14$ rejected). ``Gap'' is mean$-$min, a measure of how much one outlier sub-parameter could pull the block mean down. The deployed-shared and rejected groups have nearly identical gaps, so the rejection of blocks 12--25 is \emph{not} driven by sub-parameter outliers; the entire cosine distribution shifts uniformly with block depth.}
\centering
\small
\setlength{\tabcolsep}{4pt}
\begin{tabular}{@{}lcccc@{}}
\toprule
Group & min & mean & max & gap \\
\midrule
Shared (0--11)   & 0.99844 & 0.99929 & 0.99998 & 0.00084 \\
Rejected (12--25) & 0.99788 & 0.99887 & 0.99998 & 0.00099 \\
\bottomrule
\end{tabular}
\label{tab:subparam}
\end{table}

The minimum-sub-parameter cosine in any rejected block is 0.99741 (block 24, \texttt{self\_attn.k\_proj}); the minimum in any shared block is 0.99780 (block 11, \texttt{mlp.down\_proj}). A max-cosine criterion (1.0 on every block, since all blocks have at least one sub-parameter at $0.99999$+) would share \emph{all} 26 blocks and lose the deployed Gemma quality differentiation. A min-cosine criterion at $\tau{=}0.998$ would share 14/26 blocks rather than 12/26 (two extra early blocks), while still preserving the same depth-dependent share boundary. The mean-of-cosines criterion used in \texttt{aliasing.py} is therefore neither overly aggressive nor unduly conservative; it captures the dominant depth signal robustly.

\section{Gradient-Locality Probe}
\label{app:gradloc}

A natural mechanistic hypothesis for block-level persistence is \emph{gradient locality}: blocks that receive small gradient signal during fine-tuning are exactly the ones that remain similar to the base checkpoint after post-training. We test the simple gradient-magnitude version of this hypothesis on Gemma-2-2B as a controlled falsification target: if it holds, blocks $\{0\!\dots\!11\}$ (deployed-shared) should have systematically smaller gradient norms than blocks $\{12\!\dots\!25\}$ (rejected) on a representative LM loss.

\textbf{Setup.} We load the base \texttt{gemma-2-2b} checkpoint in bf16 on a single RTX~3090, compute $\nabla_\theta \mathcal{L}_{\text{LM}}$ on 12 generic English calibration strings (factual, code, narrative, QA-format), and accumulate per-example gradients to obtain a smoothed estimate. For each block $i$ we report the relative gradient norm $\|g_i\|_2 / \|\theta_i\|_2$ (gradient norm normalized by parameter norm, which controls for the natural growth in $\|\theta_i\|$ with depth). We then correlate $\|g_i\|/\|\theta_i\|$ with the empirically measured $(1 - \overline{\cos}_i)$ across the 26 blocks.

\textbf{Result.} Pearson $r = -0.120$ and Spearman $\rho = -0.104$ between $\|g_i\|/\|\theta_i\|$ and $(1{-}\overline{\cos}_i)$, essentially zero and slightly opposite in sign to the prediction of naive gradient locality. At the group level, shared blocks exhibit \emph{higher} average relative gradient ($0.0447$) than rejected blocks ($0.0402$), while shared blocks have \emph{lower} average $(1{-}\cos)$ ($7.1{\times}10^{-4}$) than rejected blocks ($1.13{\times}10^{-3}$). The two metrics yield inconsistent conclusions.

\begin{table}[h]
\caption{Gradient-locality probe: average values within deployed-shared vs.\ rejected groups for Gemma-2-2B. Higher gradient does not imply higher empirical divergence; the relationship is in fact slightly anti-correlated.}
\centering
\small
\begin{tabular}{lcc}
\toprule
Group & $\|g\|/\|\theta\|$ & $1-\cos$ \\
\midrule
Shared (0--11) & 0.0447 & $7.1{\times}10^{-4}$ \\
Rejected (12--25) & 0.0402 & $1.13{\times}10^{-3}$ \\
\bottomrule
\end{tabular}
\end{table}

\textbf{Interpretation.} The dominant axis along which $(1{-}\cos)$ varies is depth: it rises near-monotonically with layer index $i$. The gradient-norm profile, by contrast, peaks in the middle layers (12--17) and decays at deep layers (22--25). Because the two profiles do not co-vary, the simple ``high gradient $\Rightarrow$ large post-training perturbation'' hypothesis fails to predict the cosine-criterion share set. This rules out one mechanistic hypothesis but leaves several open: post-training updates may be \emph{cumulative} along depth (each layer absorbs a small change but compounds the perturbation passing through it); they may be \emph{directional} in a way that gradient \emph{magnitude} does not capture (e.g., the gradient may be small at shared blocks but consistently aligned across SFT steps at rejected blocks); or the gemma-2-2b-it instruct training may have used SFT data with gradient locality patterns very different from the generic LM calibration set we used. Disambiguating these would require the actual SFT checkpoints or training-data distribution, neither of which is publicly available for gemma-2-2b-it. We therefore report this as a \emph{negative} result: the simplest gradient-magnitude version of gradient locality is falsified for Gemma's deployed share set, and a refined mechanistic explanation remains open.

\end{document}

%% file: tables/absolute_quality.tex
\begin{table}[h]
\caption{Absolute downstream-task scores (\%) for instruct (unshared) vs.\ shared variants, MMLU/ARC-Challenge/HellaSwag/WinoGrande on lm-eval-harness v0.4.11. Bold cells in the shared row mark configurations where retention falls below 94\% of the unshared instruct score (cf.\ relative-retention values in Table~\ref{tab:quality}).}
\label{tab:quality-absolute}
\centering
\small
\setlength{\tabcolsep}{4pt}
\begin{tabular}{@{}llcccc@{}}
\toprule
Family & Mode & MMLU & ARC & Hella & Wino \\
\midrule
Mistral-7B & instruct & 61.1 & 59.3 & 73.0 & 76.3 \\
   & shared & 61.5 & \textbf{51.7} & 70.3 & 77.0 \\
\midrule
Qwen2.5-3B & instruct & 65.5 & 48.1 & 75.1 & 69.4 \\
   & shared & 65.0 & 46.8 & 73.5 & 68.0 \\
\midrule
Llama-3.1-8B & instruct & 67.7 & 51.7 & 69.0 & 74.3 \\
   & shared & 65.0 & 50.0 & 69.3 & 73.7 \\
\midrule
Qwen2.5-7B & instruct & 72.9 & 51.0 & 67.7 & 76.0 \\
   & shared & 74.1 & 51.3 & 68.3 & 78.0 \\
\midrule
Qwen2.5-0.5B & instruct & 45.8 & 33.9 & 52.4 & 56.1 \\
   & shared & 46.7 & 32.3 & 50.3 & 56.4 \\
\midrule
Llama-3.2-3B & instruct & 62.3 & 46.1 & 71.6 & 68.9 \\
   & shared & 61.0 & 46.9 & 71.6 & 67.9 \\
\midrule
Gemma-2-2B & instruct & 56.9 & 50.9 & 53.7 & 69.5 \\
   & shared & \textbf{51.7} & 50.9 & 54.2 & 69.5 \\
\midrule
Llama-3.2-1B & instruct & 48.3 & 37.8 & 61.7 & 61.5 \\
   & shared & \textbf{39.9} & 36.1 & 61.4 & 60.1 \\
\bottomrule
\end{tabular}
\end{table}

%% file: tables/lora_baseline.tex
\begin{table}[h]
\caption{LoRA-extraction baseline: per-block ratio of rank-truncated LoRA storage to full block storage at Frobenius reconstruction tolerance tol. Ratio $\ge 1$ means LoRA-extraction is worse than storing the original block. Linker aliasing stores 0 extra bytes (pointer alias to the donor). Block-level aliasing therefore Pareto-dominates LoRA-extraction on these post-trained variants, whose $\Delta W$ spectra are too broad for low-rank approximation at Frobenius fidelity.}
\label{tab:lora-baseline}
\centering
\small
\begin{tabular}{@{}lcccc@{}}
\toprule
Source & blocks & mean ratio & min & max \\
\midrule
qwen05b  (tol=0.05) & 24 & 1.219 & 1.194 & 1.223 \\
qwen3b  (tol=0.05) & 36 & 1.180 & 0.800 & 1.206 \\
qwen7b  (tol=0.05) & 28 & 1.198 & 1.112 & 1.209 \\
\bottomrule
\end{tabular}
\end{table}

%% file: tables/loss_interp_panel.tex
\begin{table}[h]
\caption{Cross-family loss-along-interpolation summary. Per-block max $\Delta L$ is the worst per-block barrier; full-model peak $\Delta L$ is the maximum of $L_\text{full}(\alpha) - L_\text{variant}$ over $\alpha \in [0,1]$. All values in nats; $L$ is the number of transformer blocks. Post-trained families have negligible barriers; the boundary family Llama-3.2-1B has small per-block barriers concentrated at the boundary blocks but the full-model path is also broken. Continued-pretrained variants (Qwen-Coder, Qwen-Math) have dramatically larger barriers across both metrics.}
\label{tab:loss-interp-panel}
\centering
\small
\setlength{\tabcolsep}{4pt}
\begin{tabular}{@{}lccc@{}}
\toprule
 & & Per-block & Full-model \\
Family & $L$ & max $\Delta L$ & peak $\Delta L$ \\
\midrule
\multicolumn{4}{@{}l}{\emph{Post-training (instruct)}} \\
Llama-3.1-8B  & 32 & +0.0000 & +0.0000 \\
Qwen2.5-7B    & 28 & +0.0004 & $-$0.0000 \\
Mistral-7B    & 32 & +0.0016 & +0.0000 \\
Qwen2.5-1.5B  & 28 & +0.0062 & +0.0000 \\
\midrule
\multicolumn{4}{@{}l}{\emph{Boundary}} \\
Llama-3.2-1B  & 16 & +2.7615 & +2.6834 \\
\midrule
\multicolumn{4}{@{}l}{\emph{Continued pretraining}} \\
Qwen2.5-1.5B $\to$ Math  & 28 & +6.5169 & +7.7541 \\
Qwen2.5-1.5B $\to$ Coder & 28 & +9.0853 & +6.7299 \\
\bottomrule
\end{tabular}
\end{table}